# Automatic Cattle Identification using YOLOv5 and Mosaic Augmentation: A Comparative Analysis

Rabin Dulal[1,3], Lihong Zheng[1,3], Muhammad Ashad Kabir[1,3], Shawn McGrath[1,3],
Jonathan Medway[1,3], Dave Swain[2,3], Will Swain[2,3]
[1] Charles Sturt University, Australia
Email: {rdulal, lzheng, akabir, shmcgrath, jmedway}@csu.edu.au
[2] TerraCipher Pty Ltd, Australia
Email: {dave.swain, will.swain} @terracipher.com
[3] Food Agility CRC Ltd, Australia

*Abstract*— You Only Look Once (YOLO) is a single-stage object detection model popular for real-time object detection, accuracy, and speed. This paper investigates the YOLOv5 model to identify cattle in the yards. The current solution to cattle identification includes radio-frequency identification (RFID) tags. The problem occurs when the RFID tag is lost or damaged. A biometric solution identifies the cattle and helps to assign the lost or damaged tag or replace the RFID-based system. Muzzle patterns in cattle are unique biometric solutions like a fingerprint in humans. This paper aims to present our recent research in utilizing five popular object detection models, looking at the architecture of YOLOv5, investigating the performance of eight backbones with the YOLOv5 model, and the influence of mosaic augmentation in YOLOv5 by experimental results on the available cattle muzzle images. Finally, we concluded with the excellent potential of using YOLOv5 in automatic cattle identification. Our experiments show YOLOv5 with transformer performed best with mean Average Precision (mAP)_0.5 (the average of AP when the IoU is greater than 50%) of 0.995, and mAP_0.5:0.95 (the average of AP from 50% to 95% IoU with an interval of 5%) of 0.9366. In addition, our experiments show the increase in accuracy of the model by using mosaic augmentation in all backbones used in our experiments. Moreover, we can also detect cattle with partial muzzle images.

## I. INTRODUCTION

There is a high demand for effective cattle identification for biosecurity, food supply chain, and many more. Cattle identification systems start from manual counting to automatic identification with the help of image processing. Traditional cattle identification systems such as ear tagging [1], ear notching [2], and electronic devices [3] have been used for individual identification in cattle farming. RFID tags are popular in the industry to replace traditional cattle identification systems. RFID is easy to transfer data using a wireless system. However, tag replacement is a labor-intensive method, and there is still a possibility of losing the tag, electronic device malfunction, and tag number frauding [4]. Furthermore, it has reusability issues. These are some challenges for cattle identification in livestock farm management. To overcome the shortcomings of RFID tags, advanced machine learning, and computer vision technologies are being applied in precision livestock management, including critical disease detection, vaccination, production management, tracking, health monitoring, and animal well-being monitoring [5].

With the advent of computer-vision technology, cattle visual features have gained popularity for cattle identification [6], [7]. Visual feature-based cattle identification systems are used to detect and classify different breeds or individuals based on a set of unique features. Visual cattle identification uses images of cattle body parts or whole cattle to identify. Visual feature-based methods utilize image processing and machine learning to identify cattle according to facial or body coat patterns. These visual features are fed into various classifiers (e.g., template matching and Euclidean distance), machine learning models (e.g. KNN, ANN, and SVM), and deep learning models (e.g., CNN, ResNet, and Inception) to achieve good performance [6], [8]–[10]. Iris and retinal biometric features remain unchanged over time [11], [12]. However, the implementation shows difficulty with retinal and iris image capturing [13]. Like human fingerprints, cattle muzzle print images have been used for cattle identification [14]–[16]. Muzzle print or nose print of cattle shows distinct grooves and beaded patterns, no other body parts have these types of distinct patterns. The muzzle prints of the cattle are unique biometric identifiers so that each cattle has different muzzle prints [17]. This is the motivation of using muzzle images in our research. Machine learning models are used for cattle identification. Identification accuracy heavily relates to extracted features from the face or body [6], [9], [17], [18]. SVM [19] and KNN [20] have been used for cattle IDs since 2014. Other models such as ANNs [21], [22] and decision tree, random forest (RF), and logistic regression (LR). Deep Learning (DL) models show their superior performance in applying a bunch of filters to mine powerful features deeply for cattle identification [10], [23]. DL has been seen successfully applied in object detection and tracking. As an extended application, the top listed cattle identification models are Convolution Neural Network (CNN) [24], ResNet [25], VGG [26], AlexNet [27]
Inception [28], DETR [29], and Faster R-CNN [30]. Many deep learning models are used in cattle identification [31]–[35]. However, most of them are two-stage detectors (the first stage is the extraction of object proposal, and the second stage is localization and classification) [10], [31], [32], fail to achieve high accuracy, and have not used the latest and best-performing technology in object detection.

Moreover, researchers used YOLO [36] in cattle identification due to its real-time computation and balanced accuracy and speed [31], [37]. However, they did not do any comparative study of different backbones to extract features [31], [32] and identify which backbones perform well in cattle identification. Shen et al. [31] used YOLO with Alexnet backbone and obtained an accuracy of 96.65%. In 2021, Shojaeipour et al. [32] used YOLOv3 [38] model with Resnet50 and developed a two-stage detector, and obtained an identification accuracy of 99.11%. However, YOLO is originally meant and famous for its single-stage detection due to its speed. Xu et al. [33] used RetinaFace [39] with MobileNet [40] backbone. A feature pyramid is used for multi-scale feature fusion. The major portion of improvement is the use of an optimized loss function. The additive angular margin loss function helped to enhance the performance by separating face classes. This model has improved results with an average precision of 91.3%, and an average processing time of 0.042s, and this performance is achieved even in overlapped, illuminated, and shadowed images. Nevertheless, this experiment was performed in offline mode. More recently, Li et al. [35] tested 59 different image classification models and obtained the highest accuracy of 98.7%. However, this research did not include popular object detection models like Faster R-CNN, DETR, and YOLOv5. Similar object identification models with YOLOv5 have obtained exciting results in different tasks [41]–[45]. Again, there was not much detailed discussion on YOLOv5 architecture, a comparative analysis of different backbones, or any other model optimization procedure like augmentation, etc.

We argue that a suitable solution should address the problems of using ear tags and will enhance cattle identification and object detection. Hence, the main objectives of this research are:

- To analyze the performance of five popular object detection models for cattle identification.
- To emphasize the theoretical basis of YOLOv5 architecture.
- To compare the performance of the eight backbones in YOLOv5.
- To analyze the effect of mosaic augmentation in the YOLOv5 model.
- To experiment with the achievability of cattle identification by using a partial muzzle.

This paper comprises four sections. The first section is an introduction to the research. In the first section, we also discussed how other approaches are used in cattle identification. The second section covers the methodology of our research. We have results and analysis in the third section and conclusions in the fourth section.

## II. METHODOLOGY

To address our defined objectives, we took a publicly available cattle dataset [46]. We performed data curation and labeling in the pre-processing task. Then we investigated the accuracy of five popular object detection models using the cattle dataset. We chose five models because of their performance in object detection. All five models have achieved outstanding performance on the publicly available benchmark dataset, such as COCO [47], the PASCAL VOC [48], and Imagenet [49]. We selected the best-performing model among the five. So we went into the architecture of the selected model. Then, we investigated the effect of the famous eight backbones by selecting eight networks based on their performance in COCO [47], the PASCAL VOC [48], and ImageNet [49]. The reason why we selected specific eight backbones is described in the subsection Backbone Selection. We compared the effects of mosaic augmentation in the selected backbones. Once the model was fully trained with the best-performing backbone, we used the trained model with partial muzzles for testing.

### A. Object Detection Models

A total of five object detection models were first taken into our consideration. Those models were selected based on their optimum performance in standard datasets. Faster R-CNN achieved state-of-the-art object detection accuracy in PASCAL VOC 2007, 2012, and MS COCO datasets [30]. YOLO came with single-stage object detection (there is no extraction of object proposals like in a two-stage object detection model), which outperformed state-of-the-art models in real-time detection in speed and accuracy on the PASCAL VOC 2007 dataset [36]. YOLOv3 achieved three times faster detection than SSD [50] with the same level of accuracy [38] in COCO [47]. With the great achievement of the attention model in natural language processing, attention-based models were also implemented in object detection [29], [51]. DETR is an attention-based (transformer) model that performed well in the detection and segmentation used in the COCO dataset. YOLOv5 outperformed state-of-the-art models in terms of detection accuracy and speed [52].

### B. YOLOv5 Architecture

We highlighted the architecture and loss function for YOLOv5. Figure 1 shows the YOLOv5 architecture. The YOLO model has three main components: the backbone, the neck, and the detection head. Feature extraction is done by the backbone of the Cross Stage Partial (CSP) Network (CSPNet) [54] in YOLOv5. In YOLOv5, a bottleneck block is used with CSP. The bottleneck reduces the number of parameters and mathematical calculations (matrix multiplication) [25]. Previously, DenseNet used in the previous version, YOLOv3, could solve the vanishing gradient problem, but the heavy computation makes the network very complicated, expensive, and slow. It was found that DenseNet has the gradient information duplicated in each layer. During back-propagation, the gradient in the deeper network becomes so tiny that the network cannot update the weights, which can be a contributing factor to stopping the training of the network. This problem appeared due to the skip connection to reduce the computational power. So, DenseNet has solved the problem of vanishing gradient by removing skip-connection. In DenseNet, all feature maps from the preceding layer are connected to produce the feature maps of the successive layer by connecting every layer [55].

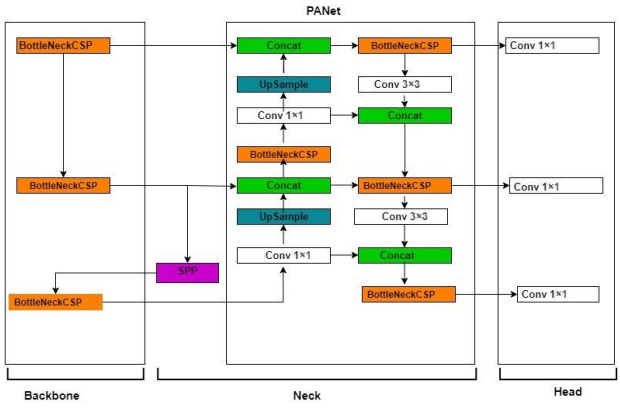

Fig. 1. The architecture of YOLOv5 [53]

To reduce this duplicated information and heavy computation, CSPNet splits the information, uses only one part in the dense layer, and combines the other part [54]. So this reintroduced the skip connection but solved the vanishing gradient problem. CSPNet makes the YOLOv5 backbone quick, fast, and intelligent enough to extract all the required features from the input images.

The job of the neck in object detection is to combine the features extracted by the backbone. Early CNNs had specific sizes and aspect ratios of images used as the input images because of the fully connected layers. These fully connected layers require predefined or fixed lengths of a vector. This job was tedious because to use a specific network, the images should be resized accordingly. This mandatory use of fixed image size came from the concept of the sliding window that is used to extract the features and combine them to produce feature maps. YOLOv5 uses Spatial Pyramid Pooling (SPP) in its neck to solve the mandatory use of fixed image size. SPP uses fixed spatial bins proportional to the input image size so that the YOLOv5 network can use images of any size, aspect ratio, and scale to produce feature pyramid vectors [56]. The combined feature pyramids are then passed to the detection head using Path Aggregation Network (PANet), which reduces the path with accurate localization information and establishes the broken path between each object proposal and object levels [57]. PANet enhanced the speed and added flexibility to the network. The detection head is the convolution network. The detection head gives information about the object with class, confidence score, location, and size [52]. A bounding box is drawn based on the information provided by the detection head.

### C. Loss Functions

YOLOv5 contains three types of losses: (*i*) classification loss, (*ii*) localization loss, and (*iii*) confidence loss.

(i) Classification loss

The *classification loss* function in YOLOv5 is called BCE-WithLogitsLoss(BCEWL). BCEWL is loss used for the multi-class classification. BCEWL is given by:

$$BCEWL = Sigmoid + BCELoss$$

And expressed as:

$$\ell c(x, y) = Lc = \{l1, c, ..., lN, c\}\top$$
$$ln, c = -wn, c[pcyn, c \cdot log\sigma(xn, c) + (1 - yn, c) \cdot log(1 - \sigma(xn, c))]$$

Where $\sigma$ is the logistic sigmoid function, c is the class number, n is the number of the sample in the batch, and $p_c$ is the weight of the positive result for class $c$. Usually, $c$ is greater than 1 for multi-label binary classification, and $c = 1$ for single-label binary classification. BCEWL suffers class imbalance which is solved by Focal Loss ($L_F$) [58] given by:

$$L_F(p) = -\alpha_t(1 - p_t)^\gamma log(p_t)$$

The weighting factor $\alpha \in [0, 1]$ and modulating factor ɤ reduces the loss contribution and extends the range where it receives low loss. This focus on the hard learning inputs and automatically down weights the easy learning inputs to make the balance in learning.

$y \in \{\pm 1\}$ is the ground truth class and $p \in [0, 1]$ is the estimated probability for y=1. Then, then

$$p_t = \begin{cases} p & when\ y = 1 \\ 1 - p & otherwise\ or\ y = 0 \end{cases}$$

The above discrete representation of focal loss is turned into continuous by Quality Focal Loss [35]:

$$QFL(\sigma) = -|y - \sigma|^\beta((1 - y)\log(1 - \sigma) + y\log(\sigma)$$

(ii) Localization loss

Complete IoU (CIoU) loss is used as localization loss which enhances the average precision and recall. The CIoU [59] is defined by:

$$CIoU = S(B, B^{gt}) + D(B, B^{gt}) + V(B, B^{gt})$$

Where S, D, and V denote the overlap area, distance, and aspect ratio.

(iii) Confidence Loss

*Confidence loss* is calculated as per the predicted bounding boxes and actual bounding boxes. Model overconfidence is minimized by using label smoothing as a regularization technique. Label smoothing reduces estimated expected calibration error (ECE). It is found that label smoothing factor $\alpha = 0.1$ improves ECE with better calibration of the model [60]. In yolov5 label smoothing is applied with $\alpha = 0.1$, then the loss with label smoothing is given by:

$$y\_true * (1.0 - label\_smoothing) + 0.5 * label\_smoothing$$

## D. Backbone Selection

After investigating the YOLOv5 model, we are interested in finding the influence of the backbone. We replaced the original backbone with seven different backbone models due to their performance on object detection and compared them with the original YOLOv5 backbone. So, a total of eight backbones is used in our experiment. In this research, we selected the backbones for specific reasons. All backbones we used in our experiments are popular for winning different competitions like ImageNet Challenge, ILSVRC classification task, and state-of-the-art models. We selected VGG because VGG came from ImageNet Challenge 2014 with first and second positions in localization and classification, respectively. VGG achieved state-of-the-art results [26]. ResNet is a widely used backbone, and its popularity came after winning the ILSVRC 2015 classification task [25]. MobileNets is popular for its lightweight and good accuracy [40]. Similarly, EfficientNet is popular for a new scaling method that scales images in all dimensions [61] are still prevalent in object detection tasks. HRNet achieved great success in object detection, semantic segmentation, and pose estimation [62]. The transformer is an attention-based model whose application in natural language processing enhanced accuracy [63]. There is an improvement in detection in YOLOv5 when a transformer is used in the head [64]. In our experiment, we used the vision transformer [51] in the backbone to facilitate feature extraction. Vision transformer is the new state-of-the-art in image classification. The use of a transformer adds emphasis to feature extraction. The last three convolutional layers of the original YOLOv5 backbone were replaced by a transformer [65]. The architecture of YOLOv5 with transformer backbone is presented in Figure 2.

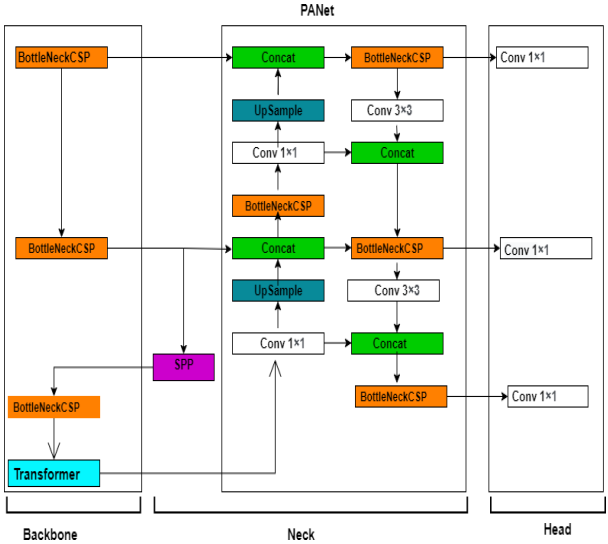

Fig. 2. The architecture of YOLOv5 with a transformer in the backbone

## E. Mosaic Augmentation

Data augmentation is used to provide diversity in training the model. Mosaic augmentation is introduced by ultralytics [66] to improve training efficiency. The ultralytics used this augmentation in YOLOv3 [67] for the first time, which was also used in YOLOv4 [68]. In mosaic augmentation, four same or different images are randomly selected, resized, cropped, and combined as a single image. After this, four bounding boxes are placed in one mosaic augmented image. The remaining part after the crop is resized and used to generate another mosaic image if they contain bounding boxes; otherwise removed. This augmentation helps to detect smaller objects because it generates four times more training data, prepares training data with different combined features that were never together in the same image to enhance learning capacity and diversity, and reduces the batch size by four times. Figure 3 shows an example of mosaic augmentation. There are parts of four different images joined together in a particular ratio to form a single image.

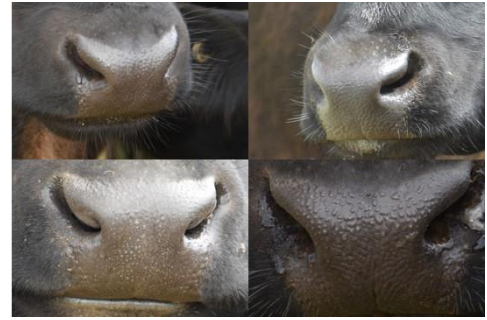

Fig. 3. An example of mosaic augmentation

## F. Partial Muzzle Identification

No prior research has focused on identifying cattle using muzzle portions. In real-world practice, if cameras are installed in a yard to capture the image, there are chances of capturing a portion of the image instead of a full muzzle image. To address the real-world scenario, we tested on portions of the muzzle. We cropped a portion of the muzzle to validate accordingly. Primarily, we took the muzzles' left, right, upper, and lower halves. The results of our testing are presented and described in the results and analysis section.

## G. Model Evaluation Metrics

The performance of the experiments is evaluated by using mean Average Precision (mAP). The map is calculated based on the precision and recall values by using the following formula:

$$Precision\ (P) = \frac{TP}{TP + FP}$$

$$Recall\ (R) = \frac{TP}{TP + FN}$$

$$AP = \int_0^1 P(R)dR, \quad \text{and} \quad mAP = \frac{1}{m}\sum_{i=1}^{m} APi$$

where m is the number of classes.

*TP*, *FP*, and *FN* denote true positive, false positive, and false negative respectively. Another important metric is the Intersection over Union (IoU). It is a common term used in object detection. If A is the detected object pixels and B is the true object pixels, then,

$$IoU(A, B) = \frac{A \cap B}{A \cup B}$$

mAP_0.5 means the average of AP when the IoU is greater than 50%, and mAP_0.5:0.95 means the average of AP from 50% to 95% IoU with an interval of 5%.

## III. RESULTS AND ANALYSIS

### A. Dataset

In our work, we used the dataset prepared by researchers from the University of New England, Armidale, Australia, via ownCloud [32]. It contains 2900 RGB images of 300 cow muzzle patterns. The image resolution is 4000 (width) × 6000 (height) in JPEG format without compression and auto lighting balance.

The dataset was numbered from 1 to 300. We cropped each image to obtain only the muzzle part, which was then labeled via roboflow [69]. The experimental dataset obtained from the labeling process contained images and corresponding labels. The dataset was divided randomly into training, validation, and testing in 70 %, 20 %, and 10 %, respectively. In addition, the data augmentation was used with a horizontal flip, vertical shear, and horizontal shear to reduce the over-fitting problem. We also used mosaic augmentation and studied the difference in performance.

### B. Models Performance

We implemented each on the prepared cattle dataset to compare the performance of the selected five object detection models. Table I shows their results. From the experimental results shown in Table I, we see the potential of using the YOLOv5 model in cattle identification. Furthermore, we used the trained YOLOv5 model's weight and used it in a video of cattle. We were able to identify cattle from the video as well. Because of the highest accuracy among the used models, we got interested in further exploring the YOLOv5 model.

TABLE I
MODEL COMPARISON

| Models | mAP 0.5 |
|---|---|
| Faster R-CNN with Resnet50 | 0.9169 |
| YOLOv3 with ResNet50 | 0.9911 |
| YOLO with Alexnet | 0.9665 |
| YOLOv5s | **0.9950** |
| DETR | 0.7720 |

### C. Effects of Different Backbones in YOLOv5

As discussed in the backbone selection subsection, we selected eight backbones. The results of the selected backbones are in Table II. Table II shows the performance of HRNet, VGG16_bn, YOLOv5s, and transformer in the original YOLOv5 model at mAP_0.5 is higher than other backbones. While considering mAP_0.5 0.95, the performance of the transformer is higher than other backbones. Figures 4, 5, and 6 provides a comparative loss of the five selected backbones. The classification loss of YOLOv5 with transformer starts with a loss smaller than 0.02, and other backbones start with a higher than 0.06. After 50 epochs, there is a significant drop in the classification loss of other backbones. By the end of the experiment, the classification loss of other backbones reaches above 0.02, but the classification loss of the transformer is lower than the starting loss. In confidence and localization loss, there is a significant drop before 50 epochs for all backbones except the transformer. While on the other hand, the transformer starts with a relatively low loss, and as the training continues, the confidence and localization loss drop steadily. The loss plot diagrams show that YOLOv5 with transformer has minimum classification loss, localization loss, and confidence loss. The loss plot diagrams also show that YOLOv5 with transformer is the best among the used backbones in our experiments.

TABLE II
PERFORMANCE IN DIFFERENT BACKBONES

| Models | mAP 0.5 | mAP 0.5:0.95 |
|---|---|---|
| YOLOv5 with HRNet | **0.9950** | 0.8605 |
| YOLOv5 with Efficientnet b1 | 0.7764 | 0.5593 |
| YOLOv5 with Efficientnet b7 | 0.7962 | 0.6673 |
| YOLOv5 with VGG16 bn | **0.9950** | 0.7475 |
| YOLOv5 with MobileNetV2 | 0.5977 | 0.5043 |
| YOLOv5 with ResNet50 | 0.7129 | 0.3989 |
| YOLOv5s with transformer | **0.9950** | **0.9366** |
| YOLOv5s | **0.9950** | 0.7992 |

Vision transformer is used as the transformer in our experiment. The transformer splits the input image into patches of equal sizes. Every patch is like a small image. Each patch is vectorized (this means the tensors are converted into vectors). A dense layer is added to each vector. The dense layer provides non-linearity in the network. A positional encoding is applied to each vector. Now each vector contains the position and content of the input image. A multi-head self-attention layer receives the output of positional encoding. This attention feature of the transformer added to the importance of the context in a certain area. Thus, the transformer backbone gives greater attention to the muzzle prints and facilitates greater feature extraction. Hence, the transformer performs better than the other seven backbones.

### D. Mosaic augmentation

The experimental results shown in Table II demonstrate the model performance with mosaic augmentation. Table III presents the performance results of YOLOv5 models with different backbones without mosaic data augmentation. In comparison with Table II and Table III, it is clear that mosaic augmentation escalates the performance.

### E. Partial Muzzle Identification

After testing partial muzzles in mosaic augmented YOLOv5 with transformer backbone, we find our model can also identify based on half-muzzle images because of mosaic augmentation. In mosaic augmentation, a unique arrangement of four patches of images is used so that the model is trained with the portion of muzzle images. Our experiment shows that the YOLOv5 with transformer cannot identify partial muzzles in the experiment performed without mosaic augmentation. Moreover, with mosaic augmentation, the model identifies with partial muzzle and bounding box can be seen with cattle ID and confidence score. Figure 7 is the original muzzle image. Figure 8 shows the left half, right half, upper half, and lower half of the muzzle shown in Figure 7. Figure 9 shows the identification of partial muzzles with a bounding box. R97 is the identified cattle. 0.87, 0.92, and 0.93 are the confidence score of the identified cattle R97.

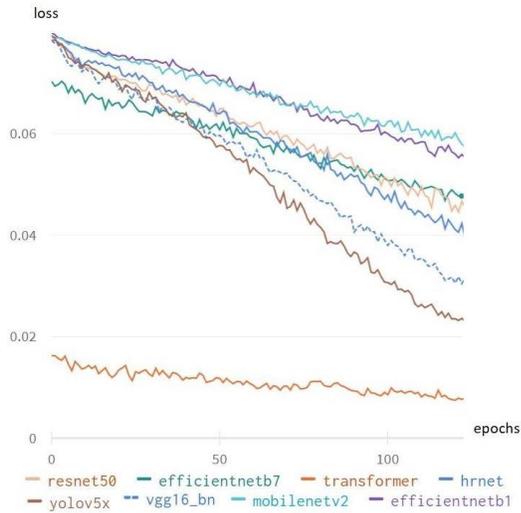

Fig. 4. Classification loss plot

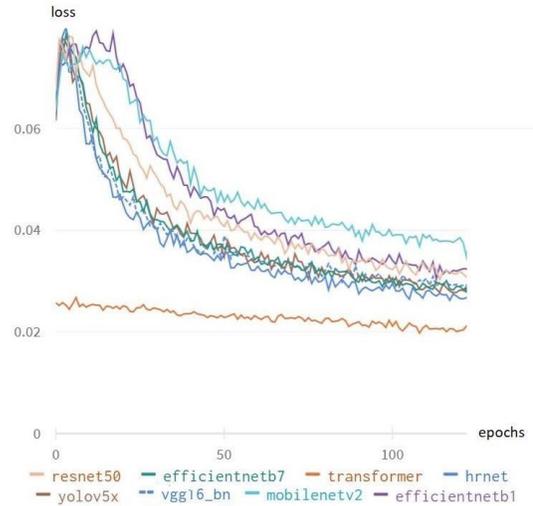

Fig. 6. Confidence loss plot

TABLE III
PERFORMANCE WITHOUT MOSAIC AUGMENTATION

| Models | mAP 0.5 | mAP 0.5:0.95 |
|---|---|---|
| YOLOv5 with hrnet | 0.3607 | 0.1530 |
| YOLOv5 with efficientnet b1 | 0.6426 | 0.2748 |
| YOLOv5 with efficientnet b7 | 0.6445 | 0.3379 |
| YOLOv5 with vgg16 bn | 0.4584 | 0.2406 |
| YOLOv5 with mobilenetV2 | 0.2750 | 0.1459 |
| YOLOv5 with resnet50 | 0.4436 | 0.1973 |
| YOLOv5s with transformer | 0.7462 | 0.3773 |
| YOLOv5s | 0.7048 | 0.3634 |

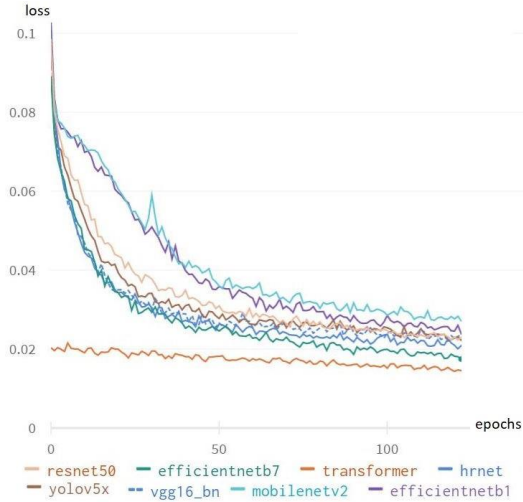

Fig. 5. Localization loss plot

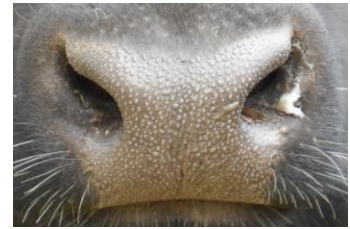

Fig. 7. Original muzzle

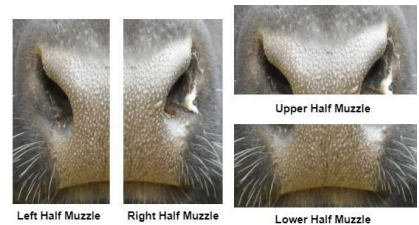

Fig. 8. Single muzzle divided into four halves

## IV. CONCLUSIONS

We have implemented and compared the popular five object detection models with the cattle dataset. We highlighted the architecture of the YOLOv5. Then, we performed experiments with eight backbones in the YOLOv5 model. Then, we inves-tigated the effects of mosaic augmentation in the same dataset with YOLOv5 in eight backbones. Based on the results of our experiments, we concluded with good potential for using YOLOv5 in automatic cattle identification. Experiments show that YOLOv5 with transformer performed best with mAP_0.5 of 0.995 and mAP_0.5:0.95 of 0.9366, which outperformed all backbones used in our experiments.

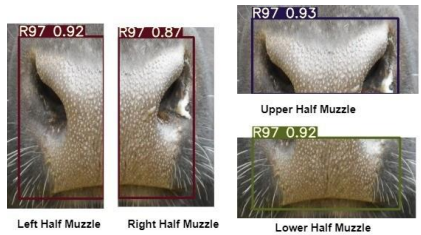

Fig. 9. Examples of partial muzzles after identification

In addition, our experiments show the increase in accuracy of the model by using mosaic augmentation in all considered backbones. Moreover, we can identify cattle with partial muzzles as well. However, mosaic augmentation 1) Improves accuracy in small object detection, 2) Works well in random objects with low object distribution, 3) Does not work in large objects because of resizing and cropping, which lose helpful information, 4) Does not work in sequential (fixed ordered), fixed location objects like handwritten text [70]. As a random crop takes the image part from any portion of the input image, there is no guarantee that the cropped part has valuable information for learning; sometimes, though a bounding box is present, it can contribute tiny. So, it may be improved by using scored crops. Each smaller portion of the image can be joined together based on the probability of contributing to the training so that valuable information will be preserved, and only highly contributing cropped patches can go through the mosaic-augmented training image.

ACKNOWLEDGMENT

This project was supported by funding from Food Agility CRC Ltd, funded under the Commonwealth Government CRC Program. The CRC Program supports industry-led collaborations between industry, researchers and the community.